\pdfoutput=1

\documentclass[11pt]{article}

\usepackage{booktabs}
\usepackage{multirow}
\usepackage{colortbl}

\usepackage[final]{acl}

\usepackage{times}
\usepackage{latexsym}
\usepackage{array}
\usepackage[T1]{fontenc}

\usepackage[utf8]{inputenc}

\usepackage{microtype}
\usepackage{inconsolata}

\usepackage{graphicx}

\usepackage{amsmath}
\usepackage{amssymb}  
\usepackage{enumitem}  

\usepackage{longtable,caption}
\definecolor{white}{HTML}{FFFFFF}
\definecolor{lightyellow}{HTML}{FFF5CC}
\definecolor{yellow}{HTML}{FFE599}
\definecolor{orange}{HTML}{FFD966}
\definecolor{strongorange}{HTML}{F4B183}

\definecolor{darkgreen}{RGB}{0,100,0}

\title{Cross-Prompt Encoder for
Low-Performing Languages}

\author{
  \textbf{Beso Mikaberidze}\textsuperscript{†}, 
  \textbf{Teimuraz Saghinadze}\textsuperscript{†},
  \textbf{Simon Ostermann}\textsuperscript{*+}, 
  \textbf{Philipp Müller}\textsuperscript{*°} 
  \vspace{0.2cm}
  \\
  \textsuperscript{†}Muskhelishvili Institute of Computational Mathematics, GTU (MICM)\\
  \textsuperscript{*}Deutsches Forschungszentrum für Künstliche Intelligenz (DFKI) 
  \\
  \textsuperscript{+} Center for European Research in Trusted AI (CERTAIN)
 \\
  \textsuperscript{°} Max Planck Institute for Intelligent Systems
  \vspace{0.2cm}
  \\
  \texttt{beso.mikaberidze@gmail.com, mueller@is.mpg.de} 
  \\
} 
\def\sibplot{
\begin{figure*}[ht]
    \centering
    \includegraphics[width=1\textwidth]{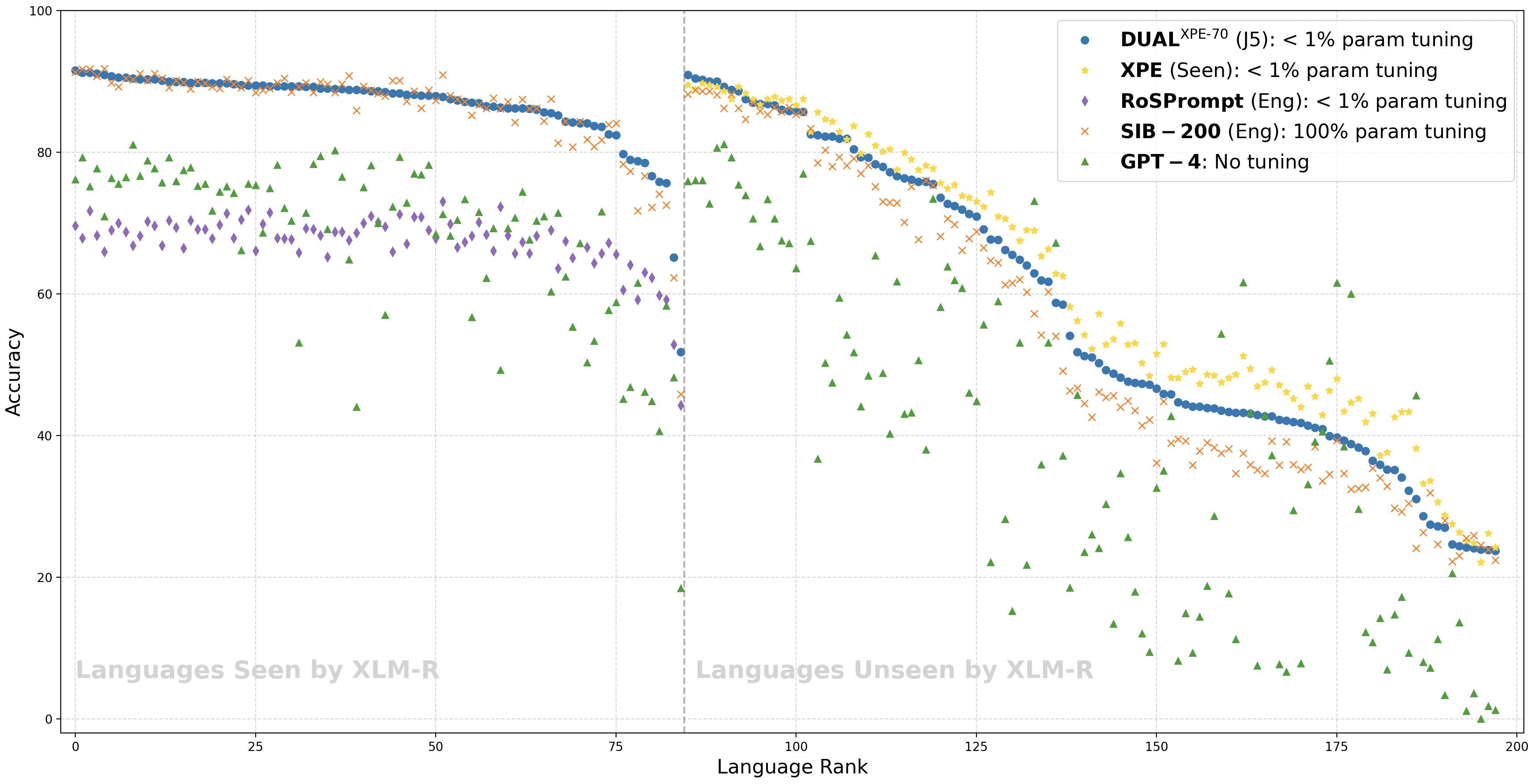}
    \caption{Comparison of different methods on the SIB-200 dataset. We group languages by whether they are seen in the pre-training corpus of XLM-R. 
    Source language groups are provided in parentheses alongside the methods, where J5 refers to the Joshi5 group. Languages are ordered by $\text{DUAL}^{\text{XPE-70}}$ (J5) performance in each group. All the methods are ZS-XLT, except for GPT-4, which is ZS prompting. It should be noted that RoSPrompts used English DBPedia14 as a source dataset 
    (a topic detection task with a different label space) and also employed language-specific verbalizers, introducing further evaluation mismatches.
    }
    \label{fig:sib200}
\end{figure*}
}

\def\sptvsxpe{
\begin{figure*}[ht]
    \centering
    \includegraphics[width=1\textwidth]{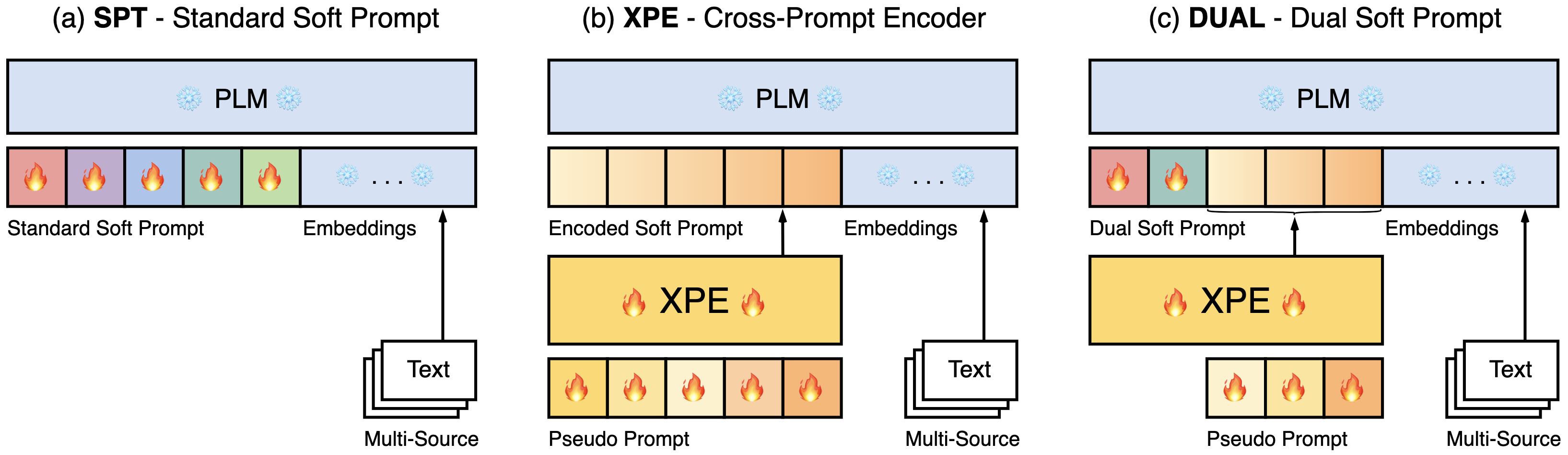}
    \caption{Architectural setup of the three methodologies during training: (a) \textbf{SPT} – Standard Soft Prompt, (b) \textbf{XPE} – Cross-Prompt Encoder, and (c) \textbf{DUAL} – Dual Soft Prompt, a hybrid combining both prior approaches.  Fire and snowflake icons indicate trainable and frozen parameters, respectively.}
    \label{fig:xpe}
\end{figure*}
}

\begin{document}
\maketitle

\begin{abstract}
Soft prompts have emerged as a powerful alternative to adapters in parameter-efficient fine-tuning (PEFT), enabling large language models (LLMs) to adapt to downstream tasks without architectural changes or parameter updates. While prior work has focused on stabilizing training via parameter interaction in small neural prompt encoders, their broader potential for transfer across languages remains unexplored. 
In this paper, we demonstrate that a prompt encoder can play a central role in improving performance on \textit{low-performing languages}—those that achieve poor accuracy even under full-model fine-tuning.
We investigate a lightweight encoder paired with multi-source training on typologically diverse languages. 
We call this architecture-training combination the Cross-Prompt Encoder (XPE), and show that it advances the capture of abstract, transferable patterns across languages.
To complement XPE, we propose a Dual Soft Prompt mechanism that combines an encoder-based prompt with a directly trained standard soft prompt. This hybrid design proves especially effective for target languages that benefit from both broadly shared structure and language-specific alignment. 
Text classification experiments with a transformer encoder (XLM-R) on the SIB-200 benchmark
reveal a consistent trade-off: XPE is most effective for low-performing languages, while hybrid variants offer broader adaptability across multilingual settings.
\end{abstract}

\section{Introduction}

Cross-lingual task transfer (XLT) seeks to leverage supervision in one or more source languages to enable task generalization to target languages. As highlighted in a recent survey on cross-lingual alignment~\cite{hammerl-etal-2024-understanding}, most existing approaches rely on supervising models in a single source language—typically English—before applying them to target languages. In contrast, multi-source training, where models are supervised on multiple labeled source languages, remains relatively underexplored~\cite{zheng-etal-2021-consistency}. Yet this setup holds significant promise: By exposing the model to multiple linguistic lenses, it encourages the learning of more robust, language-agnostic representations grounded in shared structural and semantic patterns across diverse languages.

This capability becomes especially important when transferring to \textit{low-resource} target languages—those for which even full-model fine-tuning yields suboptimal results due to a lack of data. These languages often exhibit substantial typological divergence from high-resource counterparts and lack alignment signals that typically aid transfer~\cite{lauscher-etal-2020-zero}. 
In such cases, effective zero-shot transfer remains one of the most persistent challenges in multilingual NLP, as  evidenced by the XTREME benchmark, which reveals consistently large performance gaps between English and many typologically diverse target languages across a range of tasks~\cite{xtreme}.

Building on prior works using \textit{prompt encoders} ~\cite{li-liang-2021-prefix, razdaibiedina-etal-2023-residual, liu-etal-2022-p, LIU2024208}, we 
investigate their application in the context of multi-source XLT. We explore a setup where a small, reusable neural encoder enriches
soft prompt with abstract, transferable patterns drawn from multiple, typologically diverse source languages. 
To resulting combination of architecture and training regime we refer as the Cross-Prompt Encoder (XPE).
Unlike dynamic prompt encoder-based approaches, both the encoder and its input embeddings are static at inference time, retaining the efficiency of standard soft prompting without introducing overhead.

To complement this architecture, we also propose a Dual Soft Prompt (DUAL) mechanism that adds a directly trained standard soft prompt (SPT) alongside the encoder-based prompt. 
This design enables the model to incorporate both abstract, cross-lingual patterns and more language-specific cues, offering complementary capabilities that can benefit each target language to varying degrees.

Our experiments on the SIB-200 benchmark—covering over 200 languages—demonstrate that XPE achieves strong performance on low-performing and typologically diverse target languages, while DUAL variants excel in other cases settings. Together, these findings highlight the strength of combining multilingual supervision with prompt modularity, enabling efficient XLT across a wide spectrum—from well-aligned to more challenging scenarios.

While we expect the proposed setup to generalize across architectures and tasks, this paper focuses specifically on zero-shot and fully supervised text classification with a transformer encoder backbone (XLM-R), reflecting the practical scope of this study.

Our contributions are three-fold:
\begin{enumerate}
\item 

We empirically study a parameter-efficient method that combines a soft prompt encoder with multi-source training on typologically diverse languages. This setup—referred to as the Cross-Prompt Encoder (XPE)—enhances cross-lingual task transfer (XLT) by encouraging the model to learn broadly applicable patterns. To our knowledge, this is the first study to apply prompt encoders in a hybrid, multi-source transfer setting.

\item We achieve state-of-the-art performance in both zero-shot transfer and full-data scenarios on SIB-200 text-classification task~\cite{adelani-etal-2024-sib}.  Our method outperforms zero-shot prompting models (e.g., GPT-4), prompt-based ZS-XLT methods (e.g., RoSPrompts), and full-model fine-tuning baselines (e.g., SIB-200) across a wide range of languages. It is especially effective on low-performing languages — those that remain challenging even under direct full-model fine-tuning.

\item We conduct ablation experiments to analyze the strengths of encoder-based and standard soft prompts. Our findings show that XPE is more effective in challenging low-performing scenarios, while standard soft prompts perform better when the source and target languages are closely aligned. Based on this, we introduce a Dual Soft Prompt (DUAL) mechanism that combines both, consistently yielding the best performance across multilingual settings.
\end{enumerate}

Our code and pretrained prompts are publicly available through GitHub\footnote{\url{https://github.com/bmikaberidze/XPE}}.

\section{Related Work}

With the rise of LLMs, a new paradigm of PEFT has emerged due to the size of models being fine-tuned~\cite{han2024parameterefficientfinetuninglargemodels, wang2025parameterefficientfinetuninglargemodels}. The general goal in mind is to minimize the number of parameters to be trained while enhancing model performance above in-context learning and ideally approaching the performance of full-fine tuning~\cite{liu2022few}. After validating its performance in single task / language scenarios PEFTs are often modified to work within multi-language problems~\cite{pfeiffer-etal-2020-mad, fu2022polyglot}.

\subsection{Parameter-efficient Cross-lingual Adaptation}

MAD-X~\cite{pfeiffer-etal-2020-mad}, based on adapters~\cite{pmlr-v97-houlsby19a}, is one of the first methods to be successfully extended to multilingual environments. 
Recently, LoRA~\cite{hu2021loralowrankadaptationlarge} was extended to cross-lingual scenarios using a method called FLARE~\cite{borchert2025languagefusionparameterefficientcrosslingual}.
One drawback of this method however is that all data points must be paired with their translation in the source language. 
LT-SFT~\cite{ansell-etal-2022-composable} and its more recent variation DeFT-X~\cite{simon2025deftxdenoisedsparsefinetuning} use \emph{Lottery Ticket Hypothesis} to employ masks in one case and in another SVD to obtain subnetworks that correspond to task and language separately and combine them to obtain cross-lingual transfer.

Major PEFT branches are viable for cross-lingual transfer, yet their zero-shot capabilities are constrained. A key limitation for approaches like MAD-X, LT-SFT, and DeFT-X is their dependence on language-specific components extracted through masked language modelling. These methods are inaccessible for languages with insufficient or non-existent unlabelled corpora, significantly limiting their utility in resource-scarce settings.

\subsection{Soft Prompt Tuning for Cross-lingual Tasks}

A recently emerging approach in parameter-efficient adaptation is to find prompts or prefixes using backpropagation, dubbed soft-prompts~\cite{li-liang-2021-prefix, lester-etal-2021-power}. Their success in single-task environments inspired researchers to extend soft prompts to multitask and multilingual environments.~\cite{fu2022polyglot}.

\sptvsxpe

Cross-lingual transfer can be achieved through various mechanisms, including the use of a basic soft prompt~\cite{philippy-etal-2024-soft}, a Mixture-of-Experts approach in the case of SMoP~\cite{choi-etal-2023-smop}, or the introduction of an explicit soft prompt translation mechanism in the case of MPT~\cite{10687356}. On the one hand, some researchers argue that the limited number of parameters in soft prompts enhances performance~\cite{philippy-etal-2024-soft}. However, in many other cases, some layers increase the parameter count while keeping the width of the injected prompt relatively small~\cite{10687356, choi-etal-2023-smop}.

Soft prompt based approach can be used in zero-shot scenarios; strategies are varied too, including finding a universal prompt across multiple tasks and multiple languages~\cite{fu2022polyglot} in the case of Polyprompt, tweaking loss and learning procedures, or even a template/context split fusion mechanism for UniPrompt~\cite{huang-etal-2022-zero} and RosPrompt~\cite{philippy-etal-2025-enhancing}. However, the results are difficult to compare, as they all utilize different datasets and do not necessarily employ the same method to select source and target languages. 

Out of all the data sets proposed in these articles, SIB-200 contains the largest number of languages and has additional labels, including Joshi's classification~\cite{adelani-etal-2024-sib}. This dataset lets us explore languages usually missing from a model's pretraining or those that generally underperform. Existing methods have limitations: UniPrompt cannot directly evaluate languages the model hasn't encountered, and Polyprompt, though interesting, was trained on mT5~\cite{xue-etal-2021-mt5}, making direct comparison challenging. What sets our work apart is its direct focus on underperforming languages—a gap, to our knowledge, not addressed by previous research. This distinct focus may explain why RoSPrompt underperforms in comparison to our proposed method.

\section{Methodology}
\label{ssec:ape}

To address the challenge of zero-shot cross-lingual transfer (ZS-XLT)—particularly for low-performing languages—
we investigate a reusable and parameter-efficient prompt encoder trained under a multi-source supervision regime. This setup, inspired by prior work such as P-Tuning~\cite{liu-etal-2022-p} and Multitask Prompt Tuning (MPT)~\cite{wang2023multitaskprompttuningenables}, is referred to as the Cross-Prompt Encoder (XPE) (see Figure~\ref{fig:xpe}(b)).
XPE consists of a single, reusable neural module that encodes a soft prompt using supervision from multiple typologically diverse source languages. The encoder and its inputs are shared across all languages, and the encoding process induces interactions among those input embeddings. Hence, the encoded soft prompt is able to learn abstract, language-agnostic patterns, thereby enhancing transferability, especially for low-performing and poorly aligned languages. At inference time, the encoded prompt is cached and used directly, preserving the efficiency of standard soft prompt tuning.

To complement this design, we introduce a Dual Soft Prompt (DUAL) mechanism that integrates XPE with an additional, directly trained standard soft prompt (SPT)
(see the Figure \ref{fig:xpe}(c)).
As the standard soft prompt does not involve a prompt encoder, it is expected to capture more language-specific features, which may assist in transferring to languages seen during backbone model pretraining or those closely aligned with them.
The resulting DUAL setup supports robust multilingual transfer across a broad spectrum of languages—ranging from well-aligned to low-performing ones—each may benefit to varying degrees from both components.

\subsection{Cross-Prompt Encoder (XPE)}
XPE employs a lightweight neural network that maps a small set of learnable input embeddings to outputs with the same hidden dimension as the frozen backbone model. We refer to these inputs as the \textit{pseudo prompt}, and to the network's output as the \textit{encoded soft prompt}.

Importantly, the prompt encoder and pseudo prompt are used only during training. Once training is complete, the encoder transforms the pseudo prompt into a static encoded soft prompt, which is cached and prepended at inference time—avoiding any additional computation or architectural change.

Formally, the encoder is defined as
$f_{\theta} : \mathbb{R}^{d} \to \mathbb{R}^{d}$,
where $\theta$ denotes the parameters of the encoder module and $d$ is the hidden size of the backbone Transformer, corresponding to its input embedding dimension. 
While pseudo prompt represents a matrix $n \times d$ where $n$ is the number of embeddings in the pseudo prompt, the encoder can handle only one embedding at a time. 
The resulting vectors are concatenated to form the final encoded soft prompt. The overall mapping from the full pseudo prompt to the encoded soft prompt can thus be expressed as $ F_{\theta} : \mathbb{R}^{n \times d} \to \mathbb{R}^{n \times d} $
 
The pseudo prompt $n \times d$ and encoder parameters $\theta$ are shared across all source languages. 

\subsection{Dual Soft Prompt (DUAL)}

The DUAL setup integrates XPE and standard SPT approaches, enabling the prompt to combine both encoder-based shared structure and directly learned embeddings.
Specifically, we allocate a fixed number of soft prompt embeddings to two components: the first part is dedicated to a standard soft prompt, and the second to an encoded soft prompt. These two segments are concatenated—standard first, followed by encoded—as illustrated in Figure~\ref{fig:xpe} and jointly tuned during training. 
The full soft prompt is injected at the embedding layer, while the backbone model remains frozen throughout.

Like the encoder parameters and pseudo prompt used in XPE, the standard soft prompt is shared across all source languages. So like XPE, the DUAL setup also produces a static, multilingual soft prompt, which is solely prepended to the input embeddings of the backbone model at inference time. This composite prompt preserves the overall token budget while blending both components.
We experiment with two configurations:  
$\text{DUAL}^{\text{XPE-70}}$ and  
$\text{DUAL}^{\text{XPE-30}}$.
In both variants, the numbers 70 and 30 denote the percentage of soft prompt tokens allocated to the XPE component, with the remaining tokens used for the standard SPT.

\section{Experiments}

We conduct experiments on the SIB-200 multilingual text classification benchmark, focusing on both zero-shot and fully-supervised XLT scenarios. The experiments are designed to assess the effectiveness of soft prompt tuning methods under diverse multi-source training setups, and to analyze performance across several meaningful target language groups, including the most challenging low-performing languages. All evaluations are conducted per target language, with aggregate results reported at the group level. All models are built upon the XLM-R large encoder and are compared against strong baselines, including full-model fine-tuning and zero-shot prompting. We also perform a detailed ablation study to isolate the contributions of each component.

\subsection{Experimental Setup}
Our experiments are based on the XLM-R large model, a transformer encoder pretrained on 100 languages. During training, the backbone remains frozen, and we optimize only a small set of parameters, that include soft prompt related parameters and the transformer classification head. The total number of trainable parameters remains under 0.3\% of the full model, enabling highly parameter-efficient transfer learning (PETL). 

We adopt a lightweight classification head instead of a verbalizer to ensure consistent evaluation across all \textasciitilde200 languages, including those not supported by XLM-R’s tokenizer. While this departs from a “pure” soft-prompt setup, it avoids the well-known cross-lingual bias introduced by shared verbalizers ~\cite{li-etal-2023-enhancing-cross}, as well as the limitations of language-specific verbalizers that only apply to tokenizer-covered languages ~\cite{philippy-etal-2025-enhancing}. In scenarios prioritizing plug-and-play modularity or few-shot generalization, however, a verbalizer remains a viable alternative.

We evaluate on the SIB-200 benchmark, a multilingual topic classification dataset covering 200 typologically diverse languages. 

The general setup of our experiments is a multi-source cross-lingual task transfer (XLT), conducted under two levels of supervision: zero-shot and full. In the zero-shot setting, the model is trained on labeled data from the source languages and directly applied to each target language without any target supervision. In contrast, the fully-supervised setting follows a sequential XLT setup, where the model is first tuned on the multi-source data, then further tuned on labeled data from a single target language before evaluation.

\subsection{Sources and Target Language Grouping}
\label{sec:language_grouping}
To study cross-lingual transfer dynamics under diverse conditions, we define several configurations for both source and target language sets. 
We use the following source configurations:
1. \textit{EnArZho}: A compact, high-resource, typologically diverse set comprising English, Arabic, and Mandarin Chinese. 
2. \textit{Joshi5}: A group of seven most high-resource languages classified as $5$ by Joshi et al.~\cite{joshi-etal-2020-state}.
3. \textit{Seen}: The $92$ languages that were included in the XLM-R pretraining corpus, representing the model’s seen-language space.  
Notably, each small group is a subset of bigger groups.

To better interpret transfer effectiveness, we aggregate results across four target language groups based on their relationship to XLM-R pretraining and downstream performance:
1. \textit{All /wo Joshi5}: All SIB-200 languages excluding the Joshi5 set.
2. \textit{Seen /wo Joshi5}: Consisting of only languages seen during XLM-R pretraining, excluding the Joshi5 set.
3. \textit{Unseen}: Languages not included in XLM-R's pretraining corpus.
4. \textit{Low-Performing} - We define low-performing languages as those for which XLM-R exhibits poor downstream performance, likely due to limited or ineffective representation during pretraining. 
Specifically, we identify such languages in SIB-200 by referring to full fine-tuning results on XLM-R large, reported in the original benchmark and selecting those with accuracy below $60\%$. 

We note that the \textit{Seen} and \textit{Unseen} groups form a disjoint partition of the full language set (\textit{Seen} + \textit{Unseen} = \textit{All}), and the \textit{Low-Performing} group is a strict subset of \textit{Unseen}. While this overlap is not enforced by definition, it aligns with expectations that languages unseen during pretraining tend to suffer from lower downstream performance.

For our fully-supervised experiments, we evaluate on a representative subset of 46 target languages—23 from the \textit{seen} group and 23 from the \textit{unseen} group—due to the prohibitive cost of training 200 dedicated models. Languages were selected to ensure diversity across language families, scripts, and resource levels. Although the selection process did not explicitly consider performance tiers, 11 out of the 23 unseen languages in the subset are later identified as low-performing, indicating a fair and challenging distribution.

\subsection{Methods Compared}

\newcommand{\ci}[2]{#1{\scriptsize\,\,$\pm$\,#2}} 
\newcommand{\bci}[2]{\textbf{#1}{\scriptsize\,\,$\pm$\,#2}} 

\definecolor{hl1}{RGB}{255, 249, 210}
\definecolor{hl2}{RGB}{255, 240, 170}
\definecolor{hl3}{RGB}{255, 230, 130}
\definecolor{hl4}{RGB}{255, 210, 90}

\begin{table*}[ht]
\centering
\small

\setlength{\tabcolsep}{20pt}
\renewcommand{\arraystretch}{1.2}
\begin{tabular}{@{}p{1cm}p{0.37cm}
c
>{\centering\arraybackslash}p{1.1cm}
>{\centering\arraybackslash}p{1.1cm}
>{\centering\arraybackslash}p{1.1cm}
>{\centering\arraybackslash}p{1.1cm}}

\toprule
\textbf{Target} & \textbf{Source} & \textbf{\#Source} & \textbf{SPT} & \textbf{$\text{DUAL}^{\text{XPE-30}}$} & \textbf{$\text{DUAL}^{\text{XPE-70}}$} & \textbf{XPE} \\
\midrule
\multirow{3}{*}{\textbf{LowPerf.}} 
& EnArZho & 3  & \ci{35.2}{2.4} & \cellcolor{hl1}\bci{36.5}{2.3} & \cellcolor{hl1}\ci{36.2}{2.7} & \ci{35.3}{3.2} \\
& Joshi5  & 7  & \cellcolor{hl1}\ci{36.0}{2.3} & \cellcolor{hl1}\ci{36.8}{2.5} & \cellcolor{hl2}\ci{37.3}{2.8} & \cellcolor{hl3}\bci{39.1}{2.0} \\
& Seen    & 92 & \cellcolor{hl3}\ci{39.1}{1.7} & \cellcolor{hl3}\ci{39.3}{2.3} & \cellcolor{hl3}\ci{40.3}{2.2} & \cellcolor{hl4}\bci{41.9}{1.8} \\[-1.8pt]

\midrule
\multirow{3}{*}{\textbf{Unseen}} 
& EnArZho & 3  & \ci{54.8}{2.3} & \cellcolor{hl1}\bci{56.0}{2.3} & \cellcolor{hl1}\ci{55.6}{2.6} & \ci{53.7}{2.8} \\
& Joshi5  & 7  & \cellcolor{hl1}\ci{56.0}{2.0} & \cellcolor{hl1}\ci{56.4}{2.1} & \cellcolor{hl2}\bci{57.6}{2.2} & \cellcolor{hl2}\ci{57.2}{1.8} \\
& Seen    & 92 & \cellcolor{hl3}\ci{58.9}{1.4} & \cellcolor{hl3}\ci{59.5}{1.7} & \cellcolor{hl3}\ci{60.1}{1.6} & \cellcolor{hl4}\bci{60.8}{1.4} \\[-1.8pt]

\midrule
\multirow{2}{*}{\textbf{Seen /wo J5}} 
& EnArZho & 3  & \cellcolor{hl2}\ci{84.7}{1.6} & \cellcolor{hl2}\bci{84.8}{1.8} & \cellcolor{hl2}\ci{84.6}{2.0} & \ci{82.6}{1.8} \\
& Joshi5  & 7  & \cellcolor{hl3}\ci{85.6}{1.3} & \cellcolor{hl3}\ci{85.3}{1.4} & \cellcolor{hl4}\bci{86.6}{1.1} & \cellcolor{hl2}\ci{84.5}{1.3} \\[-1.8pt]

\midrule
\multirow{2}{*}{\textbf{All /wo J5}} 
& EnArZho & 3  & \cellcolor{hl2}\ci{67.7}{2.0} & \cellcolor{hl3}\bci{68.4}{2.1} & \cellcolor{hl2}\ci{68.0}{2.3} & \ci{66.2}{2.3} \\
& Joshi5  & 7  & \cellcolor{hl3}\ci{68.7}{1.7} & \cellcolor{hl3}\ci{68.8}{1.8} & \cellcolor{hl4}\bci{70.0}{1.7} & \cellcolor{hl3}\ci{69.0}{1.6} \\[-1.8pt]

\bottomrule

\end{tabular}
\caption{ZS-XLT performance (accuracy) across different target groups. Each method was trained on different source language groups. Darker yellow indicates better performance (per target group). 
All values include 95\% two-sided confidence intervals (mean~$\pm$~half-width).
J5 refers to Joshi5 languages.}

\label{tab:spt_xpe}
\end{table*}

While our focus is on soft prompt–based transfer methods, there are relatively few established baselines for this setting on large-scale multilingual benchmarks like SIB-200.

Our main method is the Cross-Prompt Encoder (XPE), a parameter-efficient soft prompt encoding approach for multilingual transfer.
To isolate the contribution of the prompt encoder, we ablate it by removing the encoder from XPE, resulting in \textit{Standard Soft Prompt Tuning (SPT)}, which corresponds to the canonical soft prompt tuning approach widely used in prior work. This variant simultaneously serves as a baseline and a direct ablation of our method.

We additionally evaluate a hybrid setup, \textit{Dual Soft Prompting}, which combines the SPT and XPE components within a fixed prompt budget. This setup preserves the overall number of soft prompt embeddings while blending both prompt types. We experiment with two configurations: $\text{DUAL}^{\text{XPE-70}}$ and $\text{DUAL}^{\text{XPE-30}}$, where $70$ and $30$ refer to the percentage of prompt embeddings allocated to XPE.

To contextualize the performance of our approach, we compare it against several baselines, including zero-shot prompting with large language models, prompt-based transfer method using a single source language, and full-model fine-tuning on the SIB-200 benchmark. All models—except the zero-shot prompting LLM baselines—are based on the XLM-R large architecture, just like ours.

For zero-shot cross-lingual transfer (ZS-XLT), we include results from several prompting-based baselines: \textit{Phi-3.2-mini}, \textit{GPT-3.5}, and \textit{GPT-4}, each evaluated in a pure zero-shot setting without any task-specific tuning. We also compare against \textit{RoSPrompt}, a recent method that combines soft and hard prompts using English as the sole source language. 
Although evaluated on SIB-200, it is trained on DBPedia14 (a different topic classification dataset), and relies on language-specific verbalizers, making the setting not directly comparable.
Finally, we include the \textit{SIB-200 ZS-XLT} baseline, corresponding to full-model fine-tuning on a single source language (English, Arabic and Chinese), followed by zero-shot evaluation on target languages. 

We compare our fully-supervised multi-source XLT approach—based on parameter-efficient tuning—with the monolingual full-model fine-tuning baseline reported in the original SIB-200 benchmark. Both setups involve training a separate model for each target language; however, our method first performs multi-source training before adapting to each target language, enabling knowledge transfer across languages while updating less than 0.3\% (1.6M) of the model parameters. 
In contrast, the SIB-200 baseline trains all model parameters on target-language supervision, without incorporating any cross-lingual signals.

Notably, we report average results over 10 and 6 random seed runs for the zero-shot and fully supervised scenarios, respectively.

\begin{table*}[ht]
\centering
\small
\setlength{\tabcolsep}{8.9pt}

\begin{tabular}{@{}lcccccccccc@{}c}
\toprule
& \multicolumn{3}{c}{\textbf{ZS Prompting}} & \multicolumn{7}{c}{\textbf{ZS-XLT}} \\[3pt]
& \textbf{Phi-3.5} & \textbf{GPT-3.5}  & \textbf{GPT-4} 
& \multicolumn{3}{c}{\textbf{SIB-200}} & \textbf{RoS} 
& \textbf{SPT} & \textbf{$\text{DUAL}^{\text{XPE-70}}$} & \textbf{XPE} \\[1.5pt]
& -- & -- & -- & Eng & Ara & Zho & Eng & Joshi5 & Joshi5 & Seen \\
\midrule

LowPerf. 
& --    & 22.9      & 22.9  & 33.5 & 33.3 & 33.3 & --   & 36.0 & 37.3 & \textbf{41.9} \\

Unseen 
& --    & 35.7      & 39.2  & 54.0 & 54.7 & 54.3 & --   & 56.0 & 57.6 & \textbf{60.8} \\

Seen /wo J5 
& 49.02 & 55.7      & 68.1  & 86.2 & 86.5 & 86.5 & 67.3  & 85.6 & \textbf{86.6} & -- \\

All /wo J5 
& --    & 44.3      & 51.7  & 67.8 & 68.3 & 68.1 & --   & 68.7 & \textbf{70.0} & -- \\

\bottomrule
\end{tabular}
\caption{
Average accuracy across target language groups. The first header row indicates the general setup category, while the next two rows specify the individual methods and their corresponding source language(s). Baselines (Phi-3.5-mini, GPT-3.5, GPT-4, RoSPrompt, and SIB-200) are sourced from prior work, whereas SPT, XPE, and DUAL variants are our trained models. “J5” refers to the Joshi5 language group.
}
\label{tab:zs_xlt_full}
\end{table*}

\subsection{Implementation Details}

The soft prompt length is fixed at 20 virtual embeddings. Optimization is performed using Adafactor with a fixed learning rate and a cosine schedule with restarts (2 cycles).
For \textsc{XPE}, we use a learning rate of $5\text{e}{-5}$ and weight decay of $0.1$ for both the prompt encoder and the classification head.
In \textsc{SPT}, only the soft prompt is trained with a higher learning rate of $5\text{e}{-3}$ and no weight decay, while the classification head remains under the same settings as in \textsc{XPE}.
The \textsc{DUAL} configuration reflects the same settings applied to its respective components.
Training is conducted with a batch size of $32$. 
Early stopping is applied after the first learning rate cycle, with a patience of $20$ epochs for source training and $30$ for target; if early stopping does not trigger, training stops at a hard limit of $24,000$ optimization steps for source and $6,000$ for target.
All experiments are run on a single NVIDIA A100 GPU, with each training run taking approximately $30$ minutes.
We use the HuggingFace ecosystem \cite{wolf-etal-2020-transformers} to access the required artefacts, in accordance with the allowed scientific use.

\sibplot
\section{Results}

We extensively evaluate XPE and our proposed DUAL
approaches in zero-shot experiments on the SIB-200 dataset ~\cite{adelani-etal-2024-sib}.
We furthermore conduct an evaluation in the full fine-tuning scenario on the same dataset. 

\subsection{Zero-shot Experiments}

\paragraph{What mix of SPT and XPE works best?}
In \autoref{tab:spt_xpe} we present results of different soft prompt methods and different combinations of target and source languages in a zero-shot scenario.
We compare SPT with XPE as well as the two DUAL variants $\text{DUAL}^{\text{XPE-70}}$ and $\text{DUAL}^{\text{XPE-30}}$.
For the challenging set of low-performing languages, XPE achieves the best performance with 41.9 accuracy when training on all 92 seen languages 
(Welch’s unpaired t-test vs. SPT, $p=1.3\times10^{-5}$).
Decreasing the proportion of XPE embeddings in favour of SPT decreases performance in this scenario.
When training on only 7 source languages (Joshi5), the pattern of results is still the same, but when reducing the training languages to 3 (English, Arabic, Mandarin Chinese), utilising a mixture of SPT and XPE is more advantageous.
A similar general pattern can be observed when all unseen languages are used as target languages.
When training on all 92 seen languages, pure XPE reaches the best performance (60.8 accuracy).
With a reduced number of training languages, mixing XPE with SPT becomes advantageous.

When considering the less challenging transfer scenarios of seen languages as targets (excluding Joshi5), the advantage of combining SPT with XPE becomes evident. In this setup, $\text{DUAL}^{\text{XPE-70}}$ consistently attains the highest performance. It also reaches the top result with 70.0 accuracy when evaluated across all languages (except Joshi5; Welch’s unpaired t-test vs. SPT, $p=0.038$).
Per-language results are provided in Appendix~A.

\paragraph{SOTA comparison.}
Most importantly, our proposed approach $\text{DUAL}^{\text{XPE-70}}$ trained on Joshi5 outperforms all
baselines in all target language configurations.
Considering all languages except Joshi5 as target languages, $\text{DUAL}^{\text{XPE-70}}$ reaches 70.0 accuracy, followed by SPT (trained on Joshi5, 68.7 accuracy), and SIB-200 (trained on Arabic, 68.3 accuracy).
For unseen languages, including the subset of low-performing languages, we reach an even higher performance using pure XPE trained on all seen languages.
Here, XPE achieves an accuracy of 60.8, followed by $\text{DUAL}^{\text{XPE-70}}$ with 57.6 accuracy.
The best result from previous work is SIB-200 trained on Arabic with 54.7 accuracy.
These results underline that our approach is able to effectively integrate training signals from several source languages in challenging cross-lingual task transfer scenarios.
In \autoref{fig:sib200} we present a plot of per-language accuracies for different methods.
The improvements made by our proposed appraoches are highly consistent.
Only for a small number of languages our approaches are outperformed by the generally inferior GPT-4.

\begin{table}[t]
\centering
\small
\setlength{\tabcolsep}{14.3pt}
\begin{tabular}{@{}lccc@{}}
\toprule
\textbf{Target} & \textbf{\#Target} & \textbf{SIB-200} & \textbf{$\text{DUAL}^{\text{XPE-70}}$} \\
\midrule
Unseen     & 23 & 64.0              & \textbf{65.1} \\
Seen       & 23 & \textbf{88.3}     & 87.6 \\
All        & 46 & 76.1              & \textbf{76.3} \\
\bottomrule
\end{tabular}
\caption{Comparison of fully supervised methods. The direct full-model fine-tuning baseline is sourced from the SIB-200 paper. Our sequential XLT approach uses Joshi5 as the source language group and $\text{DUAL}^{\text{XPE-70}}$ as the method. Results are reported across target language groups. 
}
\label{tab:xpe70_sib200}
\end{table}

\subsection{Fully-supervised Experiments}

In addition to the zero-shot setting, we also evaluated our proposed approaches in a scenario where supervised data in the target language is available.
For computational feasibility, we evaluate on a representative subset of 23 seen and 23 unseen languages, as described in Section~\ref{sec:language_grouping}.
The results are shown in \autoref{tab:xpe70_sib200}.
Overall, our $\text{DUAL}^{\text{XPE-70}}$ approaches reaches a slight improvement over SIB-200.
When comparing the gains for unseen versus seen target languages, we see that $\text{DUAL}^{\text{XPE-70}}$ particularly excels for unseen languages, whereas it is at a slight disadvantage for seen languages.

\subsection{Additional Probing Experiment}
\label{sec:probing}

To test our hypothesis that \textsc{XPE} encodes more generalizable, language-agnostic information compared to \textsc{SPT}, we conduct a probing experiment. We convert SIB-200 into a unified language identification dataset by discarding topic labels and using language IDs as supervision. The resulting dataset is released as LID-200 on Hugging Face.\footnote{\url{https://huggingface.co/datasets/mikaberidze/lid200}}

Specifically, we train a lightweight classifier to predict language ID from hidden representations produced by frozen soft prompts and the frozen XLM-R large backbone. We probe the soft prompts trained in zero-shot experiments. For each combination of source group (Seen, Joshi5) and method (XPE, SPT), we use 10 soft prompts, each evaluated with 5 random seeds. 
This results in 50 classification scores per method and target language group, which are directly averaged to obtain a single mean score.
Lower classification accuracy corresponds to lower language identifiability, indicating that the model’s representations abstract away from language-specific features. 
Results are shown in \autoref{tab:probing}.
We observe slight, but consistently lower identifiability for \textsc{XPE} compared to \textsc{SPT}, in five out of six cases. 
Importantly, the difference is more noticeable in \textit{LowPerf.} and \textit{Unseen} target groups, highlighting that the generalization advantage of \textsc{XPE} emerges specifically in more challenging transfer scenarios. For example, on low-performing languages with Seen sources, XPE achieves 0.78 accuracy versus 0.82 for SPT. 
For the \textit{Seen~/wo~J5} target group and its superset \textit{All~/wo~J5}, the small or reversed gap is unsurprising, as XLM-R already provides strong language-specific representations for seen languages, leaving little room for either method to exert additional influence.

\begin{table}[t]
\centering
\small

\setlength{\tabcolsep}{17.3pt}
\renewcommand{\arraystretch}{1.2}
\begin{tabular}{@{}l l 
>{\centering\arraybackslash}p{0.8cm} 
>{\centering\arraybackslash}p{0.8cm} 
@{}}
\toprule

\textbf{Source} & \textbf{Target} & \textbf{XPE} & \textbf{SPT} \\
\midrule

\multirow{2}{*}{Seen}   & LowPerf.    & \textbf{0.78} & 0.82 \\
                        & Unseen      & \textbf{0.76} & 0.78 \\
\midrule
\multirow{4}{*}{Joshi5} & LowPerf.    & \textbf{0.81} & 0.83 \\
                        & Unseen      & \textbf{0.78} & 0.80 \\
                        & Seen /wo J5 & 0.91          & \textbf{0.90} \\
                        & All /wo J5  & \textbf{0.84} & 0.85 \\

\bottomrule
\end{tabular}
\caption{
Probing results measuring language identifiability. Lower values indicate more language-agnostic representations. Each score is computed across 10 soft prompts and 5 random seeds, yielding 50 values per method and target-language group. Bold highlights lower identifiability.
}
\label{tab:probing}
\end{table}

\section{Discussion}

\subsection{Generalization vs. Specialization}
\label{sec:discuss_generalization}

Our results reveal a consistent performance divide between XPE and SPT across different language types. XPE outperforms SPT on low-performing, typologically diverse target languages, suggesting that it is better suited for generalization. This likely stems from its encoder-based structure, which encourages abstraction and captures task-relevant patterns that generalize across languages.
In contrast, SPT achieves higher scores on seen languages, where alignment with the backbone pretraining data is stronger. This indicates a tendency toward specialization, where SPT embeddings memorize language-specific patterns that can be directly exploited when sufficient overlap exists between training and target data.

While \textsc{SPT} parameterizes soft prompt embeddings independently and optimizes them directly through the task loss, \textsc{XPE} computes prompt representations via a shared encoder that also mediates their optimization.
Intuitively, this design promotes generalization in two complementary ways. Because the encoder applies the same parameters to each prompt embedding, it (1) induces consistency across their representations in the forward pass and discourages position-specific patterns, and (2) mediates gradient flow through shared weights in the backward pass, coupling the updates across embeddings and smoothing gradient signals, thereby encouraging more abstract and transferable representations.

Our results show that it is beneficial to combine SPT and XPE into a DUAL configuration that is able to perform well across a wide range of source and target language scenarios.

\subsection{Language Diversity Shifts the Balance Toward Generalization}

Our experiments across multiple source language configurations demonstrate that source language diversity plays a critical role in shaping cross-lingual performance. As the number of source languages increases—from 3 to 7 to 92—the benefits of generalization become more pronounced.
Specifically, we observe that models with stronger generalization capacity (e.g., XPE-70 and full XPE) improve consistently with increasing diversity, often surpassing more specialized approaches like SPT. This pattern holds across both seen and unseen target groups, suggesting that language diversity amplifies the value of language-agnostic task representations.
Importantly, these gains occur without exceptions across all source configurations and target groupings. This consistency highlights the universal benefit of source language diversity and supports the claim that generalization becomes increasingly crucial in multilingual transfer.

\subsection{Alternative Explanation: Capacity Matching}

Although the optimization schedule remains fixed, the number of unique training samples varies across source configurations. One might therefore attribute our findings to a capacity-matching effect: smaller models (e.g., SPT) perform better with less data, while larger ones (e.g., XPE) benefit from greater diversity. However, the evidence instead points to architectural bias.
SPT consistently performs best on seen, well-aligned languages, regardless of source diversity and size. In contrast, XPE outperforms SPT on low-performing targets across all configurations. This persistent divide suggests that inductive biases—specialization versus generalization—play a more decisive role than model size or training volume.
Additionally, the consistent advantage of the DUAL setup in diverse settings suggests that combining architectural biases is more critical than model capacity alone.

\section{Conclusion}
We apply a parameter-efficient prompt encoder in a multi-source cross-lingual transfer setting. This setup, referred to as the Cross-Prompt Encoder (XPE), 
updates less than 0.3\% of model parameters while achieving substantial gains in the most challenging scenario—zero-shot transfer to low-performing languages. To further boost adaptability, we propose a Dual Soft Prompt mechanism that combines XPE with standard soft prompts, leveraging both abstract, transferable patterns and language-specific memorization. This hybrid design enables robust multilingual transfer across a wide spectrum of target languages, each benefiting to varying degrees from the complementary strengths of both components.

\section*{Limitations}

This work focuses on a single backbone model (XLM-R), which limits conclusions about the general applicability of XPE and DUAL to other architectures, such as encoder-decoder or decoder-only models. We evaluated our method on the SIB-200 dataset.
While this dataset has a large variety of languages, it is centred on a single task: multilingual topic classification. 
Further investigation is needed to assess generalization of our approach across different task types, including reasoning and language generation tasks.
Finally, while we explore multilingual transfer, cross-task—and more broadly, universal cross-task and cross-lingual—generalization, as explored in the Polyglot Prompt paper~\cite{fu-etal-2022-polyglot}, remains an open direction. 
These design choices reflect the practical scope of this study and define a foundation for broader future extensions.

\section*{Acknowledgements}
This work was partially supported by the European Union under Horizon Europe project \emph{"GAIN"} (GA \#101078950) and by the German Federal Ministry of Research, Technology and Space (BMFTR) as part of the project TRAILS (01IW24005).

\bibliography{custom}

\clearpage
\onecolumn

\appendix
\section{ZS-XLT Per-Language Results}

\newcommand{\colw}{0.89cm}
\setlength{\tabcolsep}{3pt}
\renewcommand{\arraystretch}{0.95}
\setlength\LTpre{0pt}
\setlength\LTpost{0pt}
\small
\begin{longtable}{lc
>{\centering\arraybackslash}p{\colw}
>{\centering\arraybackslash}p{\colw}
>{\centering\arraybackslash}p{\colw}
>{\centering\arraybackslash}p{\colw}|
>{\centering\arraybackslash}p{\colw}
>{\centering\arraybackslash}p{\colw}
>{\centering\arraybackslash}p{\colw}
>{\centering\arraybackslash}p{\colw}|
>{\centering\arraybackslash}p{\colw}
>{\centering\arraybackslash}p{\colw}
>{\centering\arraybackslash}p{\colw}
>{\centering\arraybackslash}p{\colw}
}
\label{tab:appendix_table}\\

\toprule
\addlinespace[5pt] 
\textbf{Language} & \textbf{Group} & \multicolumn{4}{c}{\textbf{EnArZho}} & \multicolumn{4}{c}{\textbf{Joshi5}} & \multicolumn{4}{c}{\textbf{Seen}} \\
\addlinespace[4pt] 
& & SPT & $\text{D}^{\text{XPE-30}}$ & $\text{D}^{\text{XPE-70}}$ & XPE & SPT & $\text{D}^{\text{XPE-30}}$ & $\text{D}^{\text{XPE-70}}$ & XPE & SPT & $\text{D}^{\text{XPE-30}}$ & $\text{D}^{\text{XPE-70}}$ & XPE \\
\addlinespace[1pt] 
\midrule
\endfirsthead
\toprule
\addlinespace[5pt] 
\textbf{Language} & \textbf{Group} & \multicolumn{4}{c}{\textbf{EnArZho}} & \multicolumn{4}{c}{\textbf{Joshi5}} & \multicolumn{4}{c}{\textbf{Seen}} \\
\addlinespace[4pt] 
& & SPT & $\text{D}^{\text{XPE-30}}$ & $\text{D}^{\text{XPE-70}}$ & XPE & SPT & $\text{D}^{\text{XPE-30}}$ & $\text{D}^{\text{XPE-70}}$ & XPE & SPT & $\text{D}^{\text{XPE-30}}$ & $\text{D}^{\text{XPE-70}}$ & XPE \\
\addlinespace[1pt] 
\midrule
\endhead
\bottomrule
\endfoot

\bottomrule
\caption*{\textbf{ZS-XLT performance (accuracy) per target language.} The "Language" column corresponds to language codes in SIB-200 dataset. While all listed languages collectively form the \textit{All~/wo~Joshi5} group, the "Group" column indicates which sub-target group(s) each language belongs to: 
2~$=$~\textit{Low-Performing} and \textit{Unseen}, 
1~$=$~\textit{Unseen}, 
0~$=$~\textit{Seen~/wo~Joshi5}.
Each set of four method columns corresponds to a different source language group. $\text{D}^{\text{XPE-30}}$ and $\text{D}^{\text{XPE-30}}$ and $\text{D}^{\text{XPE-70}}$ are abbreviated forms of the $\text{DUAL}$ setup.
}
\endlastfoot
ace\_Arab & 2 & 36.5 & 34.9 & 34.4 & 32.4 & 36.1 & 35.6 & 35.9 & 33.6 & 36.5 & 35.5 & 37.7 & 37.2 \\
aka\_Latn & 2 & 42.3 & 44.5 & 45.0 & 41.8 & 44.3 & 43.8 & 45.9 & 47.3 & 48.0 & 48.4 & 50.4 & 52.9 \\
arb\_Latn & 2 & 36.7 & 37.3 & 38.4 & 35.7 & 35.5 & 38.0 & 39.7 & 41.7 & 39.9 & 43.7 & 44.3 & 48.0 \\
ayr\_Latn & 2 & 41.0 & 41.2 & 40.5 & 42.6 & 41.7 & 42.9 & 42.1 & 45.8 & 45.5 & 47.0 & 47.6 & 46.1 \\
bam\_Latn & 2 & 30.6 & 34.9 & 36.9 & 33.8 & 32.5 & 35.0 & 34.1 & 41.3 & 36.4 & 38.0 & 38.7 & 43.3 \\
bem\_Latn & 2 & 45.1 & 45.0 & 45.1 & 41.1 & 45.8 & 46.4 & 47.6 & 47.0 & 52.7 & 52.8 & 53.1 & 52.8 \\
bjn\_Arab & 2 & 26.5 & 26.8 & 27.2 & 25.2 & 28.4 & 27.5 & 24.6 & 27.1 & 27.6 & 27.0 & 27.8 & 27.5 \\
bod\_Tibt & 2 & 22.6 & 25.7 & 24.6 & 25.4 & 24.5 & 24.1 & 24.4 & 25.8 & 25.6 & 24.7 & 26.1 & 26.3 \\
cjk\_Latn & 2 & 45.7 & 44.2 & 42.5 & 41.5 & 43.9 & 45.0 & 44.7 & 45.2 & 46.4 & 45.2 & 46.3 & 48.1 \\
ckb\_Arab & 2 & 28.4 & 29.6 & 30.6 & 25.0 & 29.5 & 33.8 & 31.0 & 29.5 & 36.7 & 35.2 & 37.0 & 38.2 \\
dyu\_Latn & 2 & 40.7 & 40.9 & 40.9 & 41.4 & 41.6 & 42.4 & 44.1 & 46.2 & 47.1 & 49.0 & 48.7 & 49.3 \\
dzo\_Tibt & 2 & 20.2 & 24.6 & 23.3 & 24.5 & 24.1 & 22.6 & 23.7 & 24.2 & 24.7 & 22.9 & 23.8 & 24.2 \\
ewe\_Latn & 2 & 34.8 & 36.1 & 36.7 & 39.1 & 34.7 & 36.8 & 37.8 & 43.3 & 40.9 & 40.8 & 42.9 & 41.9 \\
fon\_Latn & 2 & 31.1 & 33.7 & 35.0 & 35.4 & 33.3 & 33.9 & 36.4 & 41.6 & 37.2 & 36.7 & 39.2 & 43.1 \\
ibo\_Latn & 2 & 37.9 & 38.5 & 37.9 & 38.3 & 38.6 & 38.0 & 39.3 & 42.4 & 40.0 & 37.7 & 40.8 & 43.4 \\
kab\_Latn & 2 & 26.9 & 29.6 & 27.5 & 31.3 & 26.6 & 27.7 & 28.6 & 32.8 & 30.3 & 29.6 & 32.2 & 33.2 \\
kam\_Latn & 2 & 43.5 & 43.8 & 42.6 & 40.6 & 42.0 & 44.8 & 43.9 & 45.9 & 46.4 & 46.0 & 48.2 & 48.6 \\
kbp\_Latn & 2 & 29.0 & 33.7 & 32.0 & 34.4 & 31.3 & 33.7 & 32.2 & 39.2 & 36.5 & 36.3 & 35.4 & 43.3 \\
kik\_Latn & 2 & 40.2 & 43.1 & 43.4 & 43.6 & 42.3 & 43.2 & 43.8 & 45.1 & 46.7 & 47.3 & 47.2 & 48.5 \\
kin\_Latn & 2 & 40.1 & 39.0 & 37.5 & 39.0 & 39.4 & 40.7 & 39.9 & 45.5 & 44.2 & 43.9 & 45.4 & 46.3 \\
kmb\_Latn & 2 & 40.3 & 42.1 & 41.5 & 40.2 & 41.2 & 43.4 & 43.2 & 44.5 & 43.7 & 45.3 & 46.4 & 48.6 \\
knc\_Arab & 2 & 25.0 & 24.6 & 25.6 & 24.5 & 25.6 & 25.9 & 23.8 & 27.2 & 25.1 & 23.8 & 24.5 & 26.2 \\
lua\_Latn & 2 & 44.1 & 45.8 & 47.3 & 47.2 & 45.8 & 47.0 & 47.4 & 51.8 & 46.7 & 48.7 & 48.9 & 53.0 \\
lug\_Latn & 2 & 34.9 & 35.9 & 36.5 & 35.3 & 37.5 & 37.7 & 38.3 & 40.2 & 42.1 & 42.3 & 43.4 & 45.2 \\
min\_Arab & 2 & 29.0 & 28.7 & 27.3 & 26.4 & 28.5 & 29.3 & 27.2 & 30.0 & 30.4 & 29.0 & 31.1 & 30.6 \\
mni\_Beng & 2 & 32.5 & 34.5 & 33.3 & 30.0 & 33.6 & 34.0 & 35.2 & 31.4 & 38.0 & 36.5 & 35.6 & 37.6 \\
mos\_Latn & 2 & 38.5 & 40.4 & 40.0 & 38.3 & 39.8 & 42.0 & 42.9 & 44.6 & 41.0 & 44.0 & 44.9 & 46.9 \\
mri\_Latn & 2 & 36.7 & 38.9 & 37.9 & 39.2 & 36.0 & 38.4 & 38.8 & 44.1 & 35.1 & 38.8 & 39.2 & 44.6 \\
nqo\_Nkoo & 2 & 23.7 & 24.4 & 25.5 & 23.6 & 24.6 & 22.0 & 24.1 & 23.5 & 24.3 & 21.9 & 24.5 & 24.8 \\
nso\_Latn & 2 & 41.4 & 42.7 & 41.4 & 41.0 & 42.4 & 42.7 & 45.8 & 46.7 & 46.0 & 48.3 & 46.9 & 48.2 \\
nus\_Latn & 2 & 25.7 & 27.5 & 28.1 & 28.6 & 28.0 & 28.0 & 27.4 & 31.7 & 30.4 & 28.9 & 31.4 & 33.6 \\
run\_Latn & 2 & 38.0 & 41.1 & 39.7 & 39.8 & 38.9 & 40.9 & 41.4 & 44.0 & 44.1 & 44.4 & 46.4 & 46.9 \\
sat\_Olck & 2 & 22.6 & 25.0 & 25.2 & 23.4 & 24.3 & 22.4 & 23.9 & 23.1 & 23.0 & 20.8 & 22.2 & 22.1 \\
shn\_Mymr & 2 & 34.0 & 35.5 & 35.4 & 30.8 & 34.9 & 35.4 & 35.1 & 35.5 & 38.6 & 39.9 & 40.8 & 42.6 \\
smo\_Latn & 2 & 40.1 & 41.7 & 41.7 & 38.2 & 38.6 & 39.3 & 43.5 & 41.4 & 41.4 & 42.6 & 43.7 & 47.5 \\
sna\_Latn & 2 & 41.3 & 42.9 & 41.4 & 39.2 & 42.3 & 43.2 & 42.7 & 43.9 & 46.8 & 46.8 & 46.8 & 49.2 \\
sot\_Latn & 2 & 40.7 & 39.8 & 39.6 & 39.6 & 41.2 & 42.7 & 43.1 & 46.8 & 45.5 & 47.5 & 47.5 & 49.4 \\
ssw\_Latn & 2 & 45.7 & 45.3 & 45.8 & 39.6 & 45.9 & 48.1 & 49.2 & 45.1 & 52.1 & 53.0 & 52.5 & 52.8 \\
taq\_Latn & 2 & 40.7 & 42.5 & 41.4 & 41.7 & 40.2 & 42.4 & 43.3 & 44.5 & 44.5 & 45.6 & 46.5 & 48.1 \\
taq\_Tfng & 2 & 26.7 & 27.6 & 28.2 & 27.3 & 27.0 & 24.4 & 27.0 & 26.1 & 27.1 & 25.6 & 25.3 & 28.7 \\
tgk\_Cyrl & 2 & 39.4 & 40.8 & 39.1 & 42.2 & 40.3 & 42.7 & 43.2 & 44.5 & 44.9 & 47.2 & 48.6 & 51.2 \\
tsn\_Latn & 2 & 36.8 & 39.2 & 38.5 & 39.0 & 39.8 & 41.1 & 42.7 & 46.0 & 43.4 & 44.2 & 45.7 & 47.5 \\
tso\_Latn & 2 & 39.2 & 40.1 & 38.8 & 37.7 & 38.7 & 41.9 & 41.9 & 44.3 & 45.4 & 44.6 & 46.5 & 45.2 \\
tzm\_Tfng & 2 & 24.3 & 25.5 & 25.4 & 23.6 & 25.7 & 22.6 & 24.2 & 24.3 & 25.0 & 23.4 & 24.8 & 25.1 \\
umb\_Latn & 2 & 39.6 & 39.8 & 41.2 & 38.8 & 39.0 & 40.8 & 42.2 & 42.8 & 42.7 & 43.2 & 46.5 & 47.1 \\
yor\_Latn & 2 & 39.0 & 40.1 & 38.5 & 37.6 & 40.1 & 40.0 & 40.9 & 39.9 & 41.1 & 41.7 & 41.4 & 42.9 \\
ace\_Latn & 1 & 64.7 & 65.8 & 64.7 & 62.4 & 66.0 & 67.0 & 67.7 & 67.5 & 72.8 & 74.0 & 74.8 & 74.3 \\
acm\_Arab & 1 & 87.4 & 87.5 & 87.3 & 86.7 & 88.1 & 87.5 & 90.0 & 88.0 & 88.7 & 89.8 & 89.6 & 89.2 \\
acq\_Arab & 1 & 88.2 & 87.9 & 88.7 & 87.6 & 89.4 & 88.7 & 90.4 & 87.9 & 90.0 & 90.0 & 90.5 & 88.8 \\
aeb\_Arab & 1 & 85.1 & 85.6 & 85.5 & 84.5 & 87.5 & 86.7 & 87.5 & 85.7 & 87.9 & 87.3 & 87.7 & 88.3 \\
ajp\_Arab & 1 & 87.0 & 87.7 & 87.7 & 87.3 & 88.1 & 87.5 & 88.6 & 87.9 & 89.3 & 88.5 & 89.3 & 89.2 \\
apc\_Arab & 1 & 87.7 & 88.8 & 88.9 & 89.0 & 88.2 & 88.7 & 90.2 & 89.3 & 89.8 & 90.7 & 90.3 & 89.7 \\
ars\_Arab & 1 & 88.9 & 88.4 & 88.6 & 88.1 & 89.4 & 88.8 & 90.9 & 88.2 & 90.6 & 91.0 & 91.3 & 89.5 \\
ary\_Arab & 1 & 84.7 & 85.7 & 85.2 & 85.0 & 86.1 & 85.8 & 86.8 & 86.3 & 86.7 & 86.7 & 86.7 & 87.5 \\
arz\_Arab & 1 & 86.8 & 86.9 & 86.3 & 87.4 & 88.7 & 86.3 & 89.2 & 87.5 & 89.3 & 89.2 & 89.2 & 88.6 \\
ast\_Latn & 1 & 85.7 & 85.5 & 85.3 & 84.3 & 85.2 & 85.8 & 86.6 & 86.0 & 85.9 & 87.3 & 87.0 & 87.8 \\
awa\_Deva & 1 & 85.1 & 85.4 & 85.3 & 84.8 & 84.1 & 85.0 & 87.0 & 85.2 & 86.4 & 87.6 & 87.8 & 87.4 \\
bak\_Cyrl & 1 & 64.9 & 66.9 & 66.1 & 62.9 & 66.5 & 65.8 & 69.1 & 67.2 & 71.0 & 71.9 & 72.6 & 72.3 \\
ban\_Latn & 1 & 79.2 & 78.4 & 77.6 & 75.2 & 80.5 & 79.3 & 82.2 & 77.8 & 82.2 & 83.3 & 83.9 & 84.3 \\
bho\_Deva & 1 & 82.1 & 82.6 & 82.3 & 80.3 & 82.6 & 81.5 & 82.5 & 79.7 & 83.3 & 83.6 & 83.4 & 82.9 \\
bjn\_Latn & 1 & 76.4 & 75.3 & 75.5 & 72.7 & 78.1 & 76.4 & 79.3 & 75.3 & 79.7 & 80.5 & 81.6 & 79.8 \\
bug\_Latn & 1 & 61.6 & 63.3 & 61.2 & 62.1 & 62.4 & 62.7 & 65.5 & 67.2 & 67.0 & 68.7 & 69.4 & 69.4 \\
ceb\_Latn & 1 & 74.4 & 74.2 & 74.1 & 69.5 & 76.7 & 74.7 & 75.5 & 74.6 & 79.2 & 79.1 & 78.0 & 77.7 \\
crh\_Latn & 1 & 81.5 & 82.5 & 80.2 & 77.1 & 82.4 & 81.6 & 82.2 & 80.1 & 83.4 & 83.8 & 84.0 & 84.6 \\
dik\_Latn & 1 & 39.3 & 41.4 & 41.8 & 40.9 & 41.5 & 41.0 & 44.4 & 45.6 & 45.6 & 47.7 & 48.6 & 49.0 \\
fao\_Latn & 1 & 77.7 & 79.8 & 78.2 & 73.9 & 80.0 & 79.7 & 81.9 & 76.3 & 81.6 & 83.6 & 82.8 & 81.8 \\
fij\_Latn & 1 & 40.7 & 40.7 & 40.0 & 38.1 & 42.5 & 39.9 & 41.1 & 43.5 & 42.0 & 42.1 & 43.7 & 45.5 \\
fur\_Latn & 1 & 71.7 & 73.9 & 73.3 & 67.0 & 73.4 & 73.7 & 77.2 & 71.8 & 77.7 & 81.4 & 80.9 & 80.4 \\
fuv\_Latn & 1 & 46.4 & 46.7 & 46.0 & 44.0 & 48.1 & 48.5 & 48.7 & 48.4 & 51.7 & 51.7 & 52.8 & 53.6 \\
grn\_Latn & 1 & 61.1 & 64.6 & 64.4 & 65.4 & 62.2 & 63.7 & 66.2 & 67.6 & 65.9 & 66.6 & 69.0 & 70.6 \\
hat\_Latn & 1 & 54.7 & 57.2 & 55.9 & 50.7 & 57.5 & 57.7 & 58.7 & 56.9 & 62.6 & 62.7 & 63.5 & 62.8 \\
hne\_Deva & 1 & 83.5 & 85.0 & 84.1 & 82.5 & 84.7 & 84.6 & 85.8 & 83.3 & 86.4 & 87.0 & 87.2 & 86.6 \\
ilo\_Latn & 1 & 63.2 & 64.3 & 64.0 & 63.0 & 64.8 & 65.0 & 67.6 & 67.1 & 68.2 & 69.8 & 70.0 & 70.9 \\
kac\_Latn & 1 & 38.2 & 39.5 & 40.1 & 37.8 & 38.2 & 42.0 & 41.8 & 43.5 & 41.4 & 43.2 & 45.1 & 44.0 \\
kas\_Arab & 1 & 68.7 & 69.1 & 69.2 & 68.1 & 68.4 & 69.1 & 71.3 & 68.8 & 71.2 & 72.0 & 72.8 & 73.6 \\
kas\_Deva & 1 & 61.5 & 63.0 & 61.5 & 58.5 & 62.2 & 63.6 & 64.0 & 59.5 & 65.2 & 68.3 & 67.8 & 69.0 \\
kea\_Latn & 1 & 74.5 & 75.2 & 74.7 & 72.8 & 76.0 & 77.0 & 77.9 & 74.8 & 80.2 & 80.2 & 80.2 & 80.1 \\
knc\_Latn & 1 & 43.7 & 45.4 & 43.4 & 44.5 & 45.9 & 46.3 & 47.3 & 48.4 & 52.2 & 52.5 & 51.9 & 50.2 \\
kon\_Latn & 1 & 49.3 & 51.2 & 50.6 & 47.0 & 51.0 & 52.5 & 51.2 & 53.4 & 51.7 & 53.8 & 55.6 & 54.2 \\
lij\_Latn & 1 & 71.3 & 73.3 & 73.5 & 68.9 & 74.0 & 73.6 & 76.1 & 72.0 & 76.6 & 77.1 & 77.8 & 79.0 \\
lim\_Latn & 1 & 74.7 & 77.0 & 75.4 & 71.4 & 78.4 & 76.6 & 79.2 & 76.9 & 81.3 & 82.5 & 83.5 & 82.5 \\
lin\_Latn & 1 & 49.3 & 49.3 & 48.4 & 46.7 & 50.3 & 49.2 & 51.0 & 50.6 & 51.7 & 51.6 & 52.5 & 52.2 \\
lmo\_Latn & 1 & 71.7 & 74.1 & 73.1 & 64.9 & 74.5 & 75.4 & 76.3 & 71.4 & 77.7 & 79.9 & 79.4 & 79.9 \\
ltg\_Latn & 1 & 73.1 & 74.3 & 72.0 & 68.4 & 74.9 & 72.7 & 75.8 & 72.3 & 76.5 & 77.2 & 77.3 & 78.1 \\
ltz\_Latn & 1 & 67.0 & 70.3 & 69.7 & 65.6 & 68.7 & 69.6 & 72.7 & 69.5 & 73.7 & 74.6 & 75.0 & 74.9 \\
luo\_Latn & 1 & 42.0 & 42.6 & 40.5 & 41.3 & 41.5 & 43.0 & 44.1 & 44.3 & 46.9 & 46.8 & 47.6 & 47.3 \\
lus\_Latn & 1 & 58.1 & 58.7 & 59.3 & 56.2 & 59.6 & 59.2 & 61.9 & 60.0 & 64.6 & 66.6 & 65.5 & 65.3 \\
mag\_Deva & 1 & 83.8 & 84.6 & 84.7 & 82.8 & 84.4 & 83.9 & 85.8 & 83.7 & 86.5 & 87.5 & 87.0 & 87.5 \\
mai\_Deva & 1 & 84.3 & 84.5 & 85.1 & 83.9 & 84.9 & 84.3 & 86.8 & 83.7 & 86.8 & 86.9 & 86.6 & 86.7 \\
min\_Latn & 1 & 80.0 & 79.3 & 78.4 & 73.6 & 82.1 & 81.7 & 82.4 & 79.4 & 84.9 & 86.1 & 86.1 & 85.6 \\
mlt\_Latn & 1 & 57.2 & 60.2 & 58.4 & 55.8 & 60.0 & 60.1 & 62.9 & 61.5 & 65.9 & 67.8 & 66.4 & 68.9 \\
nya\_Latn & 1 & 47.8 & 48.3 & 48.7 & 47.0 & 48.4 & 51.3 & 51.8 & 50.7 & 52.0 & 54.6 & 54.2 & 56.2 \\
oci\_Latn & 1 & 82.1 & 84.8 & 84.0 & 81.2 & 84.1 & 83.9 & 86.0 & 84.4 & 86.1 & 86.2 & 86.5 & 87.3 \\
pag\_Latn & 1 & 69.5 & 70.2 & 70.0 & 68.8 & 70.4 & 70.1 & 70.9 & 71.1 & 71.3 & 73.3 & 72.7 & 73.0 \\
pap\_Latn & 1 & 73.8 & 76.0 & 75.7 & 71.7 & 74.4 & 75.7 & 76.6 & 75.1 & 76.8 & 77.5 & 78.1 & 77.5 \\
prs\_Arab & 1 & 89.3 & 89.2 & 88.8 & 87.9 & 88.8 & 89.1 & 90.0 & 87.8 & 90.2 & 89.9 & 90.0 & 89.5 \\
quy\_Latn & 1 & 49.4 & 51.4 & 50.5 & 51.4 & 51.2 & 50.4 & 54.1 & 54.6 & 54.4 & 56.7 & 57.4 & 58.2 \\
sag\_Latn & 1 & 42.7 & 46.6 & 44.2 & 40.6 & 42.8 & 45.1 & 47.2 & 47.4 & 46.1 & 47.2 & 47.8 & 48.4 \\
scn\_Latn & 1 & 68.3 & 70.8 & 70.1 & 66.1 & 70.4 & 71.2 & 73.6 & 69.8 & 73.3 & 74.7 & 75.0 & 75.6 \\
srd\_Latn & 1 & 72.3 & 74.5 & 72.4 & 66.3 & 74.5 & 74.1 & 75.8 & 70.2 & 77.7 & 78.0 & 78.0 & 77.5 \\
szl\_Latn & 1 & 76.8 & 78.4 & 77.0 & 72.4 & 77.9 & 77.0 & 80.4 & 76.4 & 81.6 & 82.4 & 82.8 & 83.7 \\
tat\_Cyrl & 1 & 69.0 & 70.7 & 68.9 & 64.8 & 71.5 & 71.2 & 72.4 & 70.1 & 75.0 & 76.5 & 75.8 & 75.4 \\
tgl\_Latn & 1 & 83.2 & 83.7 & 83.3 & 82.6 & 85.0 & 84.8 & 85.7 & 86.0 & 86.1 & 86.2 & 87.1 & 87.5 \\
tir\_Ethi & 1 & 52.6 & 56.8 & 58.4 & 49.3 & 53.6 & 57.3 & 58.5 & 54.4 & 59.8 & 63.9 & 62.9 & 62.5 \\
tpi\_Latn & 1 & 67.2 & 68.2 & 68.5 & 62.5 & 69.0 & 69.1 & 71.9 & 66.7 & 70.9 & 70.6 & 71.6 & 73.8 \\
tuk\_Latn & 1 & 59.8 & 59.0 & 56.5 & 53.9 & 61.9 & 60.2 & 61.7 & 55.6 & 65.3 & 65.6 & 65.4 & 66.3 \\
tum\_Latn & 1 & 42.6 & 43.5 & 44.3 & 40.9 & 43.5 & 48.2 & 46.6 & 47.8 & 49.1 & 49.0 & 50.2 & 51.5 \\
twi\_Latn & 1 & 42.0 & 45.8 & 46.7 & 43.5 & 44.8 & 44.9 & 48.2 & 49.0 & 48.8 & 51.2 & 52.8 & 55.8 \\
vec\_Latn & 1 & 76.4 & 80.0 & 78.8 & 73.7 & 79.4 & 79.2 & 81.9 & 77.1 & 82.2 & 84.2 & 84.4 & 82.9 \\
war\_Latn & 1 & 75.5 & 74.9 & 74.8 & 72.1 & 77.4 & 76.8 & 78.3 & 76.6 & 79.9 & 81.3 & 80.9 & 80.9 \\
wol\_Latn & 1 & 48.3 & 48.5 & 47.7 & 50.7 & 49.7 & 51.1 & 50.2 & 53.7 & 56.7 & 56.1 & 55.5 & 57.2 \\
yue\_Hant & 1 & 87.7 & 87.4 & 87.0 & 88.7 & 87.7 & 88.0 & 88.8 & 87.0 & 87.5 & 89.0 & 89.3 & 87.5 \\
zul\_Latn & 1 & 58.8 & 60.8 & 60.0 & 51.6 & 60.9 & 61.7 & 64.8 & 56.7 & 66.2 & 67.2 & 67.7 & 67.5 \\
afr\_Latn & 0 & 87.2 & 87.5 & 87.7 & 86.5 & 87.8 & 87.8 & 89.4 & 88.2 &  &  &  &  \\
als\_Latn & 0 & 86.7 & 87.2 & 86.9 & 86.4 & 87.9 & 87.9 & 89.2 & 87.8 &  &  &  &  \\
amh\_Ethi & 0 & 80.9 & 81.7 & 81.1 & 79.6 & 82.2 & 81.3 & 82.4 & 81.1 &  &  &  &  \\
asm\_Beng & 0 & 80.8 & 82.5 & 83.4 & 74.8 & 82.5 & 83.7 & 86.3 & 80.0 &  &  &  &  \\
azb\_Arab & 0 & 74.7 & 73.8 & 73.1 & 72.8 & 75.2 & 73.9 & 75.8 & 75.2 &  &  &  &  \\
azj\_Latn & 0 & 87.3 & 87.7 & 87.0 & 87.9 & 87.6 & 87.5 & 88.8 & 89.3 &  &  &  &  \\
bel\_Cyrl & 0 & 86.3 & 87.5 & 87.5 & 86.9 & 88.2 & 87.8 & 89.6 & 86.6 &  &  &  &  \\
ben\_Beng & 0 & 84.7 & 84.4 & 84.4 & 82.7 & 84.7 & 84.8 & 86.2 & 85.8 &  &  &  &  \\
bos\_Latn & 0 & 88.2 & 87.9 & 88.7 & 87.7 & 88.8 & 88.8 & 90.3 & 89.0 &  &  &  &  \\
bul\_Cyrl & 0 & 87.5 & 87.4 & 88.0 & 88.8 & 89.2 & 88.7 & 89.0 & 88.6 &  &  &  &  \\
cat\_Latn & 0 & 89.0 & 88.4 & 88.6 & 87.2 & 89.2 & 89.0 & 90.3 & 88.1 &  &  &  &  \\
ces\_Latn & 0 & 89.6 & 89.0 & 89.8 & 88.9 & 90.5 & 89.7 & 90.9 & 89.6 &  &  &  &  \\
cym\_Latn & 0 & 83.8 & 83.9 & 82.9 & 78.7 & 84.5 & 85.0 & 86.2 & 82.9 &  &  &  &  \\
dan\_Latn & 0 & 88.3 & 88.7 & 89.0 & 88.3 & 89.5 & 89.0 & 90.7 & 89.2 &  &  &  &  \\
ell\_Grek & 0 & 86.2 & 88.2 & 87.7 & 85.4 & 88.2 & 88.1 & 89.2 & 87.1 &  &  &  &  \\
epo\_Latn & 0 & 88.4 & 87.5 & 87.3 & 85.7 & 89.0 & 87.9 & 89.3 & 86.4 &  &  &  &  \\
est\_Latn & 0 & 89.1 & 87.5 & 87.7 & 84.2 & 89.1 & 88.6 & 89.7 & 88.6 &  &  &  &  \\
eus\_Latn & 0 & 86.7 & 85.8 & 85.9 & 84.2 & 88.4 & 86.8 & 89.3 & 85.7 &  &  &  &  \\
fin\_Latn & 0 & 87.5 & 87.8 & 87.1 & 85.9 & 88.7 & 88.2 & 89.3 & 87.7 &  &  &  &  \\
gaz\_Latn & 0 & 47.9 & 49.3 & 46.7 & 39.6 & 50.2 & 51.7 & 51.8 & 46.6 &  &  &  &  \\
gla\_Latn & 0 & 75.8 & 76.5 & 75.6 & 59.9 & 77.1 & 77.4 & 78.7 & 69.6 &  &  &  &  \\
gle\_Latn & 0 & 82.3 & 83.3 & 80.9 & 73.1 & 83.3 & 84.1 & 85.2 & 79.8 &  &  &  &  \\
glg\_Latn & 0 & 88.7 & 88.3 & 87.8 & 88.2 & 88.5 & 88.7 & 88.9 & 88.3 &  &  &  &  \\
guj\_Gujr & 0 & 84.4 & 84.8 & 84.2 & 82.5 & 84.6 & 84.2 & 86.2 & 84.5 &  &  &  &  \\
hau\_Latn & 0 & 76.7 & 76.6 & 72.3 & 64.1 & 78.5 & 78.4 & 79.7 & 73.5 &  &  &  &  \\
heb\_Hebr & 0 & 86.2 & 85.9 & 85.9 & 85.0 & 86.3 & 86.5 & 87.1 & 85.9 &  &  &  &  \\
hin\_Deva & 0 & 87.7 & 87.6 & 87.7 & 85.7 & 87.7 & 87.3 & 88.3 & 87.4 &  &  &  &  \\
hrv\_Latn & 0 & 89.6 & 89.5 & 89.5 & 89.2 & 90.2 & 89.9 & 91.1 & 89.2 &  &  &  &  \\
hun\_Latn & 0 & 87.9 & 87.7 & 88.2 & 87.9 & 88.6 & 88.8 & 88.7 & 87.7 &  &  &  &  \\
hye\_Armn & 0 & 86.8 & 86.9 & 86.8 & 84.0 & 87.1 & 86.8 & 88.6 & 86.8 &  &  &  &  \\
ind\_Latn & 0 & 90.1 & 90.1 & 90.3 & 89.2 & 90.7 & 90.0 & 91.2 & 90.3 &  &  &  &  \\
isl\_Latn & 0 & 87.3 & 86.6 & 86.8 & 86.9 & 88.8 & 87.6 & 89.3 & 88.9 &  &  &  &  \\
ita\_Latn & 0 & 89.1 & 88.9 & 89.5 & 88.8 & 89.3 & 90.3 & 90.4 & 89.6 &  &  &  &  \\
jav\_Latn & 0 & 83.4 & 83.3 & 82.3 & 80.2 & 84.3 & 84.2 & 84.1 & 82.2 &  &  &  &  \\
kan\_Knda & 0 & 85.5 & 85.6 & 87.0 & 83.3 & 86.9 & 86.7 & 89.0 & 86.6 &  &  &  &  \\
kat\_Geor & 0 & 87.5 & 88.5 & 88.4 & 86.8 & 88.5 & 88.4 & 89.5 & 89.5 &  &  &  &  \\
kaz\_Cyrl & 0 & 86.8 & 87.4 & 87.0 & 85.8 & 87.9 & 87.6 & 87.8 & 87.9 &  &  &  &  \\
khk\_Cyrl & 0 & 83.4 & 84.1 & 84.6 & 83.2 & 85.0 & 84.0 & 86.1 & 84.8 &  &  &  &  \\
khm\_Khmr & 0 & 84.5 & 86.2 & 85.6 & 84.5 & 85.6 & 85.5 & 87.3 & 86.6 &  &  &  &  \\
kir\_Cyrl & 0 & 85.6 & 86.1 & 86.1 & 87.4 & 87.6 & 85.9 & 88.5 & 87.5 &  &  &  &  \\
kmr\_Latn & 0 & 76.3 & 76.9 & 76.8 & 69.4 & 77.8 & 77.3 & 78.9 & 74.6 &  &  &  &  \\
kor\_Hang & 0 & 86.6 & 86.9 & 86.1 & 87.1 & 87.2 & 86.8 & 88.0 & 88.2 &  &  &  &  \\
lao\_Laoo & 0 & 85.1 & 86.9 & 86.7 & 86.6 & 86.6 & 85.6 & 88.8 & 88.7 &  &  &  &  \\
lit\_Latn & 0 & 86.9 & 86.5 & 87.1 & 85.5 & 86.8 & 87.0 & 89.4 & 88.3 &  &  &  &  \\
lvs\_Latn & 0 & 88.5 & 87.3 & 87.5 & 87.8 & 88.3 & 89.2 & 89.8 & 87.7 &  &  &  &  \\
mal\_Mlym & 0 & 85.4 & 86.6 & 85.2 & 82.5 & 87.5 & 85.6 & 86.9 & 85.6 &  &  &  &  \\
mar\_Deva & 0 & 87.4 & 87.0 & 85.8 & 82.4 & 87.4 & 86.3 & 88.1 & 85.1 &  &  &  &  \\
mkd\_Cyrl & 0 & 85.9 & 86.5 & 86.3 & 85.4 & 86.8 & 86.9 & 88.1 & 85.9 &  &  &  &  \\
mya\_Mymr & 0 & 83.5 & 85.1 & 85.0 & 83.1 & 85.3 & 84.9 & 86.5 & 84.5 &  &  &  &  \\
nld\_Latn & 0 & 88.9 & 89.2 & 89.2 & 87.7 & 89.2 & 89.4 & 90.0 & 89.5 &  &  &  &  \\
nno\_Latn & 0 & 88.6 & 88.8 & 88.2 & 86.8 & 88.9 & 88.7 & 90.5 & 88.8 &  &  &  &  \\
nob\_Latn & 0 & 88.1 & 87.1 & 87.2 & 87.5 & 88.1 & 87.6 & 88.6 & 87.8 &  &  &  &  \\
npi\_Deva & 0 & 86.2 & 86.1 & 86.6 & 84.4 & 86.7 & 87.3 & 87.5 & 85.2 &  &  &  &  \\
ory\_Orya & 0 & 81.8 & 81.4 & 81.8 & 79.5 & 83.0 & 83.4 & 84.1 & 80.8 &  &  &  &  \\
pan\_Guru & 0 & 83.6 & 83.2 & 84.2 & 80.8 & 84.1 & 83.3 & 85.6 & 83.1 &  &  &  &  \\
pbt\_Arab & 0 & 82.0 & 81.8 & 81.6 & 79.6 & 82.2 & 82.3 & 84.2 & 80.8 &  &  &  &  \\
pes\_Arab & 0 & 89.9 & 89.7 & 89.6 & 90.1 & 89.9 & 89.7 & 90.5 & 89.8 &  &  &  &  \\
plt\_Latn & 0 & 74.6 & 75.8 & 73.4 & 66.8 & 77.8 & 77.0 & 78.5 & 71.1 &  &  &  &  \\
pol\_Latn & 0 & 88.7 & 88.7 & 89.2 & 88.0 & 90.0 & 89.1 & 89.9 & 87.7 &  &  &  &  \\
por\_Latn & 0 & 88.5 & 88.6 & 87.7 & 87.5 & 89.1 & 88.4 & 89.8 & 87.3 &  &  &  &  \\
ron\_Latn & 0 & 87.9 & 87.6 & 87.6 & 87.0 & 88.9 & 89.0 & 89.7 & 87.5 &  &  &  &  \\
rus\_Cyrl & 0 & 87.8 & 87.7 & 88.0 & 88.7 & 88.9 & 88.0 & 88.9 & 87.9 &  &  &  &  \\
san\_Deva & 0 & 81.7 & 81.8 & 81.2 & 78.7 & 81.8 & 81.3 & 83.7 & 80.9 &  &  &  &  \\
sin\_Sinh & 0 & 83.4 & 83.8 & 83.4 & 81.7 & 84.8 & 83.8 & 86.0 & 82.6 &  &  &  &  \\
slk\_Latn & 0 & 88.6 & 87.8 & 88.1 & 88.4 & 89.0 & 88.7 & 89.7 & 89.4 &  &  &  &  \\
slv\_Latn & 0 & 86.6 & 85.2 & 86.1 & 85.8 & 87.4 & 87.1 & 88.3 & 85.9 &  &  &  &  \\
snd\_Arab & 0 & 83.3 & 82.9 & 83.3 & 81.4 & 83.7 & 82.8 & 84.3 & 84.1 &  &  &  &  \\
som\_Latn & 0 & 72.5 & 74.1 & 72.2 & 67.7 & 74.3 & 74.1 & 75.6 & 71.1 &  &  &  &  \\
srp\_Cyrl & 0 & 87.7 & 87.7 & 88.8 & 87.3 & 89.0 & 87.9 & 89.9 & 89.0 &  &  &  &  \\
sun\_Latn & 0 & 84.4 & 84.3 & 84.9 & 81.3 & 86.4 & 85.0 & 87.0 & 83.9 &  &  &  &  \\
swe\_Latn & 0 & 88.1 & 88.7 & 88.6 & 87.2 & 88.8 & 88.5 & 89.4 & 88.4 &  &  &  &  \\
swh\_Latn & 0 & 79.6 & 80.3 & 79.9 & 79.1 & 81.2 & 81.2 & 83.6 & 81.6 &  &  &  &  \\
tam\_Taml & 0 & 84.8 & 84.5 & 84.5 & 83.3 & 85.3 & 85.6 & 86.4 & 85.4 &  &  &  &  \\
tel\_Telu & 0 & 85.7 & 86.0 & 87.1 & 85.2 & 86.2 & 86.0 & 87.9 & 86.9 &  &  &  &  \\
tha\_Thai & 0 & 87.6 & 87.7 & 87.8 & 89.8 & 89.1 & 88.2 & 89.8 & 89.1 &  &  &  &  \\
tur\_Latn & 0 & 88.0 & 87.7 & 88.5 & 87.5 & 88.9 & 88.4 & 89.3 & 87.5 &  &  &  &  \\
uig\_Arab & 0 & 81.1 & 81.8 & 81.5 & 79.1 & 82.8 & 81.4 & 82.5 & 80.5 &  &  &  &  \\
ukr\_Cyrl & 0 & 90.0 & 88.9 & 89.8 & 89.3 & 90.7 & 90.7 & 91.2 & 90.3 &  &  &  &  \\
urd\_Arab & 0 & 86.2 & 86.3 & 86.3 & 86.6 & 87.0 & 86.2 & 88.0 & 87.5 &  &  &  &  \\
uzn\_Latn & 0 & 84.3 & 84.1 & 83.0 & 81.3 & 85.0 & 84.7 & 85.5 & 83.1 &  &  &  &  \\
vie\_Latn & 0 & 89.8 & 89.7 & 89.2 & 90.4 & 90.1 & 90.2 & 90.1 & 90.0 &  &  &  &  \\
xho\_Latn & 0 & 60.4 & 62.5 & 62.4 & 51.5 & 62.5 & 62.9 & 65.1 & 58.5 &  &  &  &  \\
ydd\_Hebr & 0 & 76.3 & 73.1 & 71.7 & 65.6 & 75.7 & 76.6 & 76.6 & 69.8 &  &  &  &  \\
zho\_Hant & 0 & 89.9 & 89.3 & 89.0 & 89.2 & 89.5 & 90.0 & 90.3 & 88.7 &  &  &  &  \\
zsm\_Latn & 0 & 90.5 & 89.4 & 89.2 & 88.3 & 90.3 & 89.8 & 91.5 & 89.9 &  &  &  &  \\

\end{longtable}

\twocolumn
\clearpage

\end{document}